\documentclass[letterpaper, 10 pt, conference]{ieeeconf}  % Comment this line out if you need a4paper

\usepackage{tikz}
\usepackage[abs]{overpic}
\usepackage[normalem]{ulem}

\usepackage[symbol]{footmisc}

\usetikzlibrary{fadings,patterns}

\usepackage{blindtext}
\usepackage{colortbl}
\usepackage{graphicx}
\usepackage{subcaption}
\usepackage{arydshln}
\usepackage[acronym]{glossaries}
\usepackage{bm}
\usepackage{multirow} 
\usepackage{booktabs}

\usepackage{cite}
\usepackage{amsmath}
\usepackage{amssymb}
\usepackage{hyperref}

\newcommand{\best}{\cellcolor{tablered}}
\newcommand{\sbest}{\cellcolor{orange}}
\newcommand{\tbest}{\cellcolor{yellow}}

\definecolor{yellow}{rgb}{1, 1, 0.7}
\definecolor{orange}{rgb}{1, 0.85, 0.7}
\definecolor{tablered}{rgb}{1, 0.7, 0.7}
\definecolor{red}{rgb}{1, 0, 0}

\definecolor{myblue}{RGB}{0, 40, 139}

\definecolor{myred}{RGB}{137, 26, 15}

\definecolor{myyellow}{RGB}{167, 96, 37}

\IEEEoverridecommandlockouts                              % This command is only needed if 
\title{\LARGE \bf
SING3R-SLAM: Submap-based Indoor Monocular Gaussian SLAM\\
with 3D Reconstruction Priors}

\author{
    Kunyi Li$^{1, 3}$ \quad
    Michael Niemeyer$^{2}$ \quad
    Sen Wang$^{1, 3}$ \quad
    Stefano Gasperini$^{1, 3, 4}$ \quad\\
    Nassir Navab$^{1, 3}$ \quad
    Federico Tombari$^{1, 2}$ \vspace{0.4em} \\
    {\normalsize $^1$Technical University of Munich} \quad
    {\normalsize $^2$Google} \quad
    {\normalsize $^3$Munich Center for Machine Learning} \quad
    {\normalsize $^4$VisualAIs} \quad\\
    \vspace*{-8mm}
}

\begin{document}

%%%%%%%%% TEASER
\vspace{-5mm}
\twocolumn[{%
    \renewcommand\twocolumn[1][]{#1}%
    \maketitle
    \thispagestyle{empty}
    \begin{center}
        \includegraphics[width=0.95\linewidth]{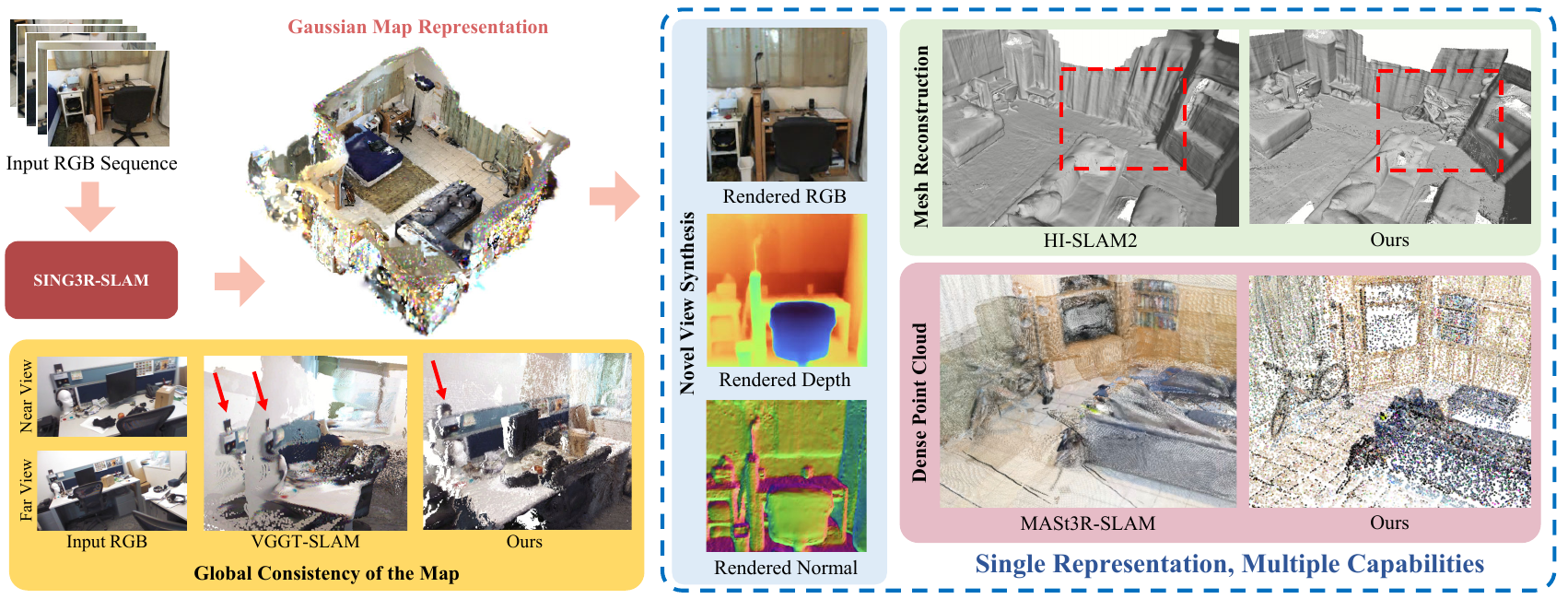}
        % \vspace{-2mm}

        \captionof{figure}{\textbf{SING3R-SLAM} is a monocular indoor SLAM system that builds a globally consistent Gaussian map for versatile downstream tasks. 
        \textbf{Left-Top:} Gaussian-based map representation that maintains multi-view consistency via differentiable rendering. 
        \textbf{Left-Bottom:} Comparison of map's global consistency with VGGT-SLAM \cite{maggio2025vggt}. 
        \textbf{Right:} Examples of downstream tasks enabled by the Gaussian map.}
        \label{fig:teaser}
    \end{center}
}]

% \maketitle
\thispagestyle{empty}
\pagestyle{empty}

\begin{abstract}
Recent advances in dense 3D reconstruction have demonstrated strong capability in accurately capturing local geometry. However, extending these methods to incremental global reconstruction, as required in SLAM systems, remains challenging. Without explicit modeling of global geometric consistency, existing approaches often suffer from accumulated drift, scale inconsistency, and suboptimal local geometry.
To address these issues, we propose SING3R-SLAM, a globally consistent Gaussian-based monocular indoor SLAM framework. Our approach represents the scene with a Global Gaussian Map that serves as a persistent, differentiable memory, incorporates local geometric reconstruction via submap-level global alignment, and leverages global map's consistency to further refine local geometry. This design enables efficient and versatile 3D mapping for multiple downstream applications.
Extensive experiments show that SING3R-SLAM achieves state-of-the-art performance in pose estimation, 3D reconstruction, and novel view rendering. It improves pose accuracy by over 10\%, produces finer and more detailed geometry, and maintains a compact and memory-efficient global representation on real-world datasets.
\end{abstract}    
\section{Introduction}
\label{sec:intro}

Visual simultaneous localization and mapping (vSLAM) lies at the core of modern robotics and augmented reality applications, enabling systems to perceive and understand their surroundings. With advances in hardware and algorithmic design, robust and accurate visual SLAM \cite{mur2015orb, teed2021droid, li2024dns, matsuki2024gaussian, li20254d} has become increasingly feasible. In particular, indoor monocular SLAM is of great importance, as it hosts a wide range of downstream applications, such as 3D scene understanding \cite{qin2024langsplat, liang2024supergseg, alegret2025gala} and navigation \cite{lei2025gaussnav}. However, achieving both precise camera pose estimation and globally consistent dense reconstruction solely from monocular inputs remains difficult. 

Previous methods that rely on 2D priors, such as optical flow \cite{horn1981determining, teed2021droid} or feature matching \cite{mur2015orb, leroy2024grounding}, achieve remarkable tracking performance and provide stable pose estimation. However, their focus is largely limited to camera localization, rather than constructing a globally consistent 3D representation that can support downstream tasks. 
Recent advances in large-scale 3D pretraining, such as DUSt3R \cite{wang2024dust3r} and MASt3R \cite{leroy2024grounding}, provide strong 3D priors. Extensions including Spann3R \cite{wang20253d}, CUT3R \cite{wang2025continuous}, and VGGT \cite{wang2025vggt} improve geometric consistency, but are typically restricted to short frame windows, limiting their ability to handle long sequences with global coherence. Sequential variants, such as SLAM3R \cite{liu2025slam3r}, MASt3R-SLAM \cite{murai2025mast3r}, and VGGT-SLAM \cite{maggio2025vggt}, further extend these priors to incremental SLAM pipelines using feature matching or submap-based strategies. 
Overall, while these SLAM frameworks leverage 3D priors for long-sequence reconstruction and pose estimation, the priors are often geometrically imprecise and remain unoptimized. Moreover, without a global memory to enforce geometric consistency, they produce locally misaligned submaps and degraded global structure over long sequences, as shown in the bottom-left of Fig.~\ref{fig:teaser}.

More recent SLAM methods \cite{keetha2024splatam, matsuki2024gaussian, li20254d, zhang2024hi} adopt 3D Gaussian Splatting (3DGS) \cite{kerbl20233d} as their map representation. Unlike previous traditional methods which use point-based maps, Gaussian maps are compact, globally consistent, and enable joint optimization of camera poses and scene geometry via differentiable rendering without complex feature matching. Their flexible and dense representation also makes them highly versatile for downstream tasks and further extensions, establishing them as the emerging standard in modern dense SLAM systems.
However, most existing systems rely on RGB-D input, limiting their applicability. RGB-only variants, such as Splat-SLAM \cite{sandstrom2025splat} and HI-SLAM2 \cite{zhang2024hi}, depend on pretrained depth and normal estimators and external tracking modules, which often produce scale-inconsistent or inaccurate depth and disentangle pose from scene reconstruction. Consequently, reconstructed maps can be geometrically imprecise or incomplete, and the use of multiple pretrained models increases computational overhead.

To overcome these limitations, we propose SING3R-SLAM, a purely monocular indoor SLAM system that requires no depth input and maintains a Global Gaussian Map as a persistent, differentiable memory. By incrementally aligning prior-based local reconstructions and jointly refining poses and geometry via differentiable rendering, our method achieves improved global geometric consistency. The result, as shown in Fig.~\ref{fig:teaser}, is a compact, dense, and globally consistent map, which not only stabilizes pose estimation but also supports high-quality reconstruction, novel view synthesis, and other downstream tasks such as 3D scene understanding.
In summary, our main contributions are as follows:
\begin{itemize}
\item We reconstruct local submaps with 3D priors and align them into a Gaussian map which acts as a global memory, offering a robust and efficient initialization for subsequent optimization.
\item We perform per frame geometry refinement via differentiable bundle adjustment, correcting inherit geometry errors from 3D priors and further enhancing the global consistency of the map.
\item We perform submap-level pose graph optimization, propagating the updated poses to the Gaussian map for further frame-level global optimization.
\end{itemize}
\section{Related Work}
\label{sec:related}

\subsection{3D Reconstruction and SLAM}
Recent advances \cite{wang2024dust3r, leroy2024grounding, wang2025continuous, wang2025vggt, wang20253d} in 3D reconstruction priors have significantly improved monocular geometry estimation and inspired new directions for visual SLAM. DUSt3R \cite{wang2024dust3r} and its successor MASt3R \cite{leroy2024grounding} pioneer two-view 3D reconstruction, achieving impressive geometric accuracy by leveraging large-scale 3D datasets and correspondence-based priors. However, as they operate on only two input views, these methods require additional global alignment to handle long sequences and maintain geometric consistency.
To overcome this limitation, Spann3R \cite{wang20253d} introduces a memory bank to extend reconstruction from two to multiple views. Despite improved temporal consistency, it remains constrained to low-resolution inputs (224×224) and only performs well for up to five frames. Moreover, the reconstructed point maps often exhibit scale inconsistencies between frames.
CUT3R \cite{wang2025continuous} and VGGT \cite{wang2025vggt} further extend multi-view reconstruction and achieve better geometric stability. Both methods still experience severe drift when applied to longer sequences.

Building on these priors, SLAM3R \cite{liu2025slam3r} is the first system to integrate DUSt3R \cite{wang2024dust3r} into a sequential framework, yet its overall performance remains limited. Similarly, MASt3R-SLAM \cite{murai2025mast3r} performs continuous two-view alignment via feature matching between adjacent frames, but its reliance on frame-by-frame matching makes it inefficient and prone to drift over long trajectories. VGGT-SLAM \cite{maggio2025vggt} reconstructs submaps from multiple frames and applies pose graph optimization on the $SL(4)$ manifold; while this improves global alignment, it primarily focuses on pose optimization rather than global geometry reconstruction, leaving the overall geometry imprecise. ViSTA-SLAM \cite{zhang2025vista} introduces Symmetric Two-view Association (STA) with shared weights for tracking, but like previous methods, it outputs unrefined point maps from the reconstruction network, without improving global map consistency.

Existing 3D reconstruction-based SLAM systems focus on pose estimation but leave geometric errors uncorrected. Without a global memory, they suffer from locally misaligned submaps and degraded global structure. In contrast, SING3R-SLAM builds a globally consistent Gaussian map, jointly refining poses and geometry to reduce drift and produce a compact, versatile representation.

\begin{figure*}[ht]
\includegraphics[width=15.5cm]{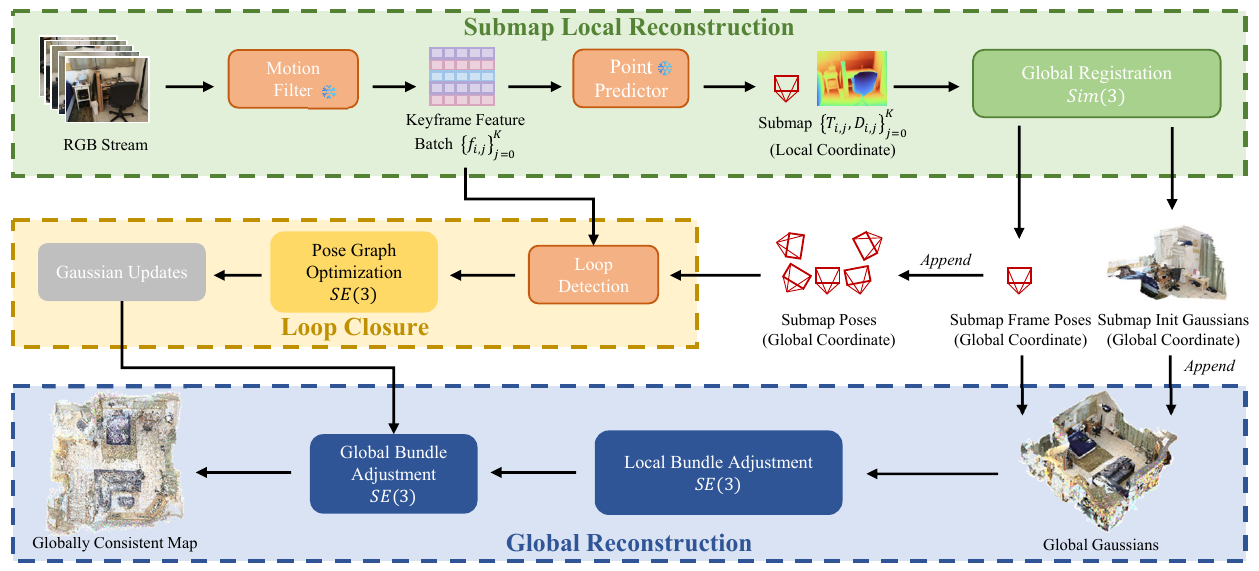}
\centering
\vspace{-2mm}
\caption{\textbf{Overview.} Our system comprises three main components: \textbf{Submap Local Reconstruction} (top), \textbf{Global Reconstruction} (bottom), and \textbf{Loop Closure} (middle). We adopt a Gaussian-based map representation as it maintains multi-view consistency via differentiable rendering and supports versatile downstream tasks. 
Our system first predicts point-based submaps from input RGB images using a monocular geometry estimator, which are then aligned and fused into a global Gaussian map. 
Differentiable bundle adjustment jointly optimizes the Gaussian map and camera poses to refine geometry and improve global consistency. 
While submap-level pose graph optimization roughly corrects loop-induced drift, global bundle adjustment further performs per-frame refinement for more accurate pose and geometry estimation.
}
\vspace{-2mm}
\label{fig:pipeline}
\end{figure*}

\subsection{3D Gaussians-based SLAM}
Unlike point-based representations \cite{wang2024dust3r, wang2025vggt}, Gaussian maps \cite{kerbl20233d} support differentiable rendering and efficient global optimization, making them well-suited for dense SLAM.
Since the training of Gaussian models typically requires registered point clouds as input, most existing Gaussian SLAM systems \cite{keetha2024splatam, yugay2023gaussian, matsuki2024gaussian} are designed for RGB-D settings. SplaTAM \cite{keetha2024splatam} and Gaussian Splatting SLAM \cite{matsuki2024gaussian} are representative examples that utilize depth input to initialize and update Gaussian parameters in real time, achieving accurate pose tracking and high-quality reconstruction but limiting applicability to depth-equipped sensors.

More recent methods \cite{li2025monogs++, sandstrom2025splat, zhang2024hi} extend Gaussian SLAM to RGB-only input. Splat-SLAM \cite{sandstrom2025splat} and HI-SLAM2 \cite{zhang2024hi} employ off-the-shelf depth and normal estimators to infer per-view geometry and use external tracking modules (e.g., DROID-SLAM \cite{teed2021droid}) for camera pose estimation. However, the predicted depth lacks scale information and global consistency, leading to cumulative drift over time. Moreover, relying on multiple pretrained modules results in fragmented pipelines with increased computational overhead.

While existing methods often rely on depth sensors or separate modules for geometry and pose estimation leading to scale ambiguities and drift, our SING3R-SLAM uses a single module to leverage local 3D priors as initialization, aligning them with a global Gaussian map to achieve improved pose and geometry reconstruction.

\section{Method}

Our method leverages a single 3D reconstruction module to first provide local geometric priors, which are then aligned and integrated into a global Gaussian map, enabling joint optimization of poses and geometry and producing a better scene representation.

\subsection{Submap Local Reconstruction}
\label{sec:tracker}
\subsubsection{Keyframe Selection}
Our system takes a sequence of RGB images $\{C_t\}_{t=0}^{N}$ as input and estimates the scene representation $\mathcal{M}$ together with the corresponding camera poses $\{T_t\}_{t=0}^{N} \in SE(3)$. To efficiently process long sequences, we first introduce a motion filter for keyframe selection. Unlike previous SLAM systems ~\cite{zhang2024hi, murai2025mast3r, maggio2025vggt} reply on 2D feature matching or optical flow, our method leverages the geometry-aware features from  the monocular geometry estimator ~\cite{wang2024dust3r, wang2025continuous}.
Given an input image $C_t \in \mathbb{R}^{H \times W \times 3}$, the encoder of the off-the-shelf monocular geometry estimator extracts patch-level features $f_t \in \mathbb{R}^{\frac{H}{16} \times \frac{W}{16} \times 1024}$, which are later used for point map and camera pose prediction. For each incoming frame, we measure its motion relative to the latest keyframe. Specifically, we flatten the patch features and compute their similarity as:
\begin{equation}
{r} = \frac{1}{N} \sum_{i=1}^{N} \mathbf{1}\big[ \max_j \langle f_t^{(i)}, f_k^{(j)} \rangle > \beta \big],
\label{eq:similarity}
\end{equation}
where $\mathbf{1[\cdot]}$ is an indicator function, $f_t^{(i)}$ is i-th feature patch of the current frame, $f_k^{(j)}$ is j-th feature patch of the latest keyframe, and $\beta=0.7$ is a similarity threshold.
If the overlap ratio $r$ is lower that a threshold indicating substantial viewpoint or scene-motion change, prompting us to select the current frame as a new keyframe.

\subsubsection{Local Geometry Prediction}
The geometry feature $f_t$ is now used to predict the local geometry and construct the local submaps. We denote the keyframe images and features as $\{C_{i,j}, f_{i,j}\}_{j=0}^{K} \in G_i$, where $G_i$ represents the $i$-th submap, $i = 0, \dots, \lfloor N/K \rfloor$, and $K$ is the number of frames per submap. To ensure temporal continuity, we introduce an overlapping frame between consecutive submaps such that the last frame $C_{i-1,K}$ of $G_{i-1}$ is identical to the first frame $C_{i,0}$ of $G_i$.
Each frame is then decoded by the monocular geometry estimator $\mathcal{D}$ to predict dense point maps and corresponding camera poses, which together constitute the submap representation:
\vspace{-1mm}
\begin{equation}
    \{X_{i,j}^{\text{self}}\}_{j=0}^K, \{T_{i,j}^{i-th}\}_{j=0}^K = \mathcal{D}(\{f_{i,j}\}_{j=0}^K),
    % \vspace{-1mm}
\end{equation}
where $\{X_{i,j}^{{\text{self}}}\}_{j=0}^{K} \in G_i$ is the point map in self coordinates, and $\{T_{i,j}^{i{-th}}\}_{j=0}^{K} \in G_i$ represents the corresponding camera poses in the $i$-th submap local coordinates. Depth maps can be further inferred from local point maps as $\{\tilde{D}_{i,j}\}_{j=0}^{K}$.

\subsection{Global Reconstruction}
\label{mapper}
We employ a global Gaussian map as a compact and differentiable scene representation, which models the scene with significantly fewer primitives than dense point maps while preserving geometric structure and supporting versatile downstream tasks. The local submaps are aligned to the coordinate system of the global Gaussian map and used to initialize the following Gaussian representation. Then we jointly optimize the Gaussians and camera poses to improve the geometry.

\subsubsection{Map Representation} 
Each 3D Gaussian $\mathcal{G}_{k} = \{{\mu}_{i, k}, \Sigma_{i, k}, \alpha_{i, k}, \mathbf{c}_{i, k}\}$ consists of a mean position ${\mu}_{i, k} \in \mathbb{R}^3$, a covariance matrix $\Sigma_{i, k} = R_{i, k} S_{i, k} S_{i, k}^{\top} R_{i, k}^{\top}$, $R_{i, k} \in SO(3), S_{i, k}=[s^0_k, s^1_k, s^2_k] \in \mathbb{R}^{3}$, an opacity $\alpha_{i, k}$ and a color $\mathbf{c}_{i, k}$. The first subscript $i$ indicates the $i$-th submap from which this Gaussian was initialized.
The Gaussian map $\mathcal{M} = \{\mathcal{G}_k\}_{k=1}^{N_g}$ serves as a global, differentiable scene representation. Given a view with the camera pose $T_{i,j}^{\text{world}}$, through a standard Gaussian Splatting \cite{kerbl20233d, zhang2024rade} process, we can render the corresponding color map $\hat{C}_{i,j}$, depth map $\hat{D}_{i,j}$, normal map $\hat{N}_{i,j}$ and silhouette map $\mathcal{A}_{i,j}$.

\subsubsection{Submap Global Registration}
The geometry of each newly predicted submap is defined in local coordinates and needs to be aligned to world coordinates. We transfer $\{T_{i,j}^{i{-th}}\}_{j=0}^{K}$ by:
\begin{equation}
\begin{aligned}
& T_{i,j}^{{\text{world}}}=T_{i-1,K}^{{\text{world}}} \, \big(T_{i,0}^{i{-th}}\big)^{-1} \, T_{i,j}^{i{-th}}, \\
\end{aligned}
\label{eq:submap_align}
\end{equation}
where $T_{i-1,K}^{{\text{world}}} \in G_{i-1}$ is the world-coordinate pose of the last frame in the previous submap. 

In contrast to VGGT-SLAM \cite{maggio2025vggt}, which performs $SL(4)$ optimization to jointly correct the alignment, scale and projection errors only in submap-level, we resolve the submap-level alignment and scale ambiguity using $SE(3)$ transformation and delegate frame-level refinement to the Gaussian optimization stage. 
Given a view with camera pose $T_{i,j}^{\text{world}}$ from newly submap, we first fix existing Gaussian map and render color $\hat{C}_{i,j}$ and depth $\hat{D}_{i,j}$ map. Then we optimize the camera pose $T_{i,j}^{\text{world}}$ by minimizing:
\begin{equation}
    \min_{T_{i,j}^{\text{world}}} \mathcal{L}_C + \lambda_{scaleD}\mathcal{L}_{scaleD},
\label{eq:pose_refine}
\end{equation}
where $\mathcal{L}_C=\| \mathcal{A}_{i,j}C_{i,j} - \mathcal{A}_{i,j}\hat{C}_{i,j}\|_1$ and $\mathcal{A}_{i,j}$ is rendered silhouette map used to ignore invalid pixels. We apply the scale-invariant depth loss to account for the potential scale inaccuracies:
\begin{equation}
\begin{aligned}
\mathcal{L}_{scaleD} = &
\sum \Big( \log \mathcal{A}_{i,j}\hat{D}_{i,j} - \log \mathcal{A}_{i,j} \tilde{D}_{i,j} \Big)^2 - \\
& \left(
\sum \Big( \log \mathcal{A}_{i,j}\hat{D}_{i,j} - \log \mathcal{A}_{i,j} \tilde{D}_{i,j} \Big)
\right)^2.
\end{aligned}
\end{equation}
Then, we correct the depth by:
\begin{equation}
    D_{i,j} = \exp\big({\log(\mathcal{A}_{i,j}\hat{D}_{i,j}) - \log(\mathcal{A}_{i,j}\tilde{D}_{i,j})}\big)\tilde{D}_{i,j}. 
\label{eq:updated_pointmap}
\end{equation}
Given camera intrinsics, the updated depth $D_{i,j}$ and pose $T_{i,j}^{{\text{world}}}$ are then used to initialize new Gaussians \cite{keetha2024splatam}.

\subsubsection{Differentiable Bundle Adjustment} 
Instead of relying on feature matching \cite{murai2025mast3r}, we perform dense differentiable bundle adjustment where camera poses and scene parameters are optimized online in a fully differentiable manner. Given a sliding window $\mathcal{V}$ with $W$ frames, we render color $\hat{C}_{i,j}$, depth $\hat{D}_{i,j}$ and normal $\hat{N}_{i,j}$ map from Gaussian map and jointly update Gaussians and poses by:
\begin{equation}
    \min_{\mathcal{M},T^{\text{world}}} \sum_{\{T^{\text{world}}\}\in\mathcal{V}} \mathcal{L}_{pho} + \lambda_{D}\mathcal{L}_{D} + \lambda_{N}\mathcal{L}_{N} + \lambda_{DN}\mathcal{L}_{DN} + \lambda_{S}\mathcal{L}_{S},
\label{eq:map_update}
\end{equation}
where $\mathcal{L}_{pho}=\| C_{i,j} - \hat{C}_{i,j}\|_1+SSIM( C_{i,j}, \hat{C}_{i,j})$ is the photometric loss \cite{kerbl20233d}. $\mathcal{L}_{D}=\| 1/D_{i,j} - 1/\hat{D}_{i,j}\|_1$ is the inverted depth loss.  $\mathcal{L}_{N}=(1 - \hat{N}_{i,j} \cdot \bar{N}_{i,j})$ is the rendered normal loss and the depth-normal ${\bar{N}}_{i,j}$ is converted from rendered depth $\hat{D}_{i,j}$. $\mathcal{L}_{DN}=(1-{N}_{i,j} \cdot \bar{N}_{i,j})$ is the depth-normal loss, and the pseudo ground truth normal map ${N}_{i,j}$ is converted from ${D}_{i,j}$. Finally, the scale loss $\mathcal{L}_{S}=\sum_{j=0}^{2} (s^j_k-\bar{s}_k)$ is used to prevent artifacts due to excessively slender Gaussians, and $\bar{s_i}$ is the mean scale value.
Together, these losses enable dense rendering-based bundle adjustment, improving global geometric consistency and correcting errors introduced by the initial priors.

\subsection{Loop Closure}
\label{sec:loop}
We further introduce a loop closure mechanism to suppress long-term drift.
Unlike previous methods \cite{maggio2025vggt, zhang2024hi, murai2025mast3r} which rely on either extra feature correspondence or bag-of-words, our approach leverages the patch-level features in Eq.\ref{eq:similarity} and globally consistent Gaussian map as a reference to perform loop detection.

\begin{table*}[t]
\centering
\resizebox{0.7\linewidth}{!}{

\begin{tabular}{lccccccccc}

\toprule
Method & \textbf{Avg.} & &chess & fire & heads & office & pumpkin & redkitchen & stairs \\ 

\midrule
CUT3R \cite{wang2025continuous} & 47.7 & &74.3 & 22.6 & 36.3 & 66.4 & 54.6 & 38.1 & 41.3 \\
MASt3R-SLAM \cite{murai2025mast3r} & 6.6 & &6.3 & 4.6 & 2.9 & 10.3 & \sbest 11.2 & 7.4 & \tbest 3.2 \\
VGGT-SLAM \cite{maggio2025vggt} & 6.7 && \sbest 3.6 & \sbest 2.8 & \best 1.8 & 10.3 & 13.3 & 5.8 & 9.3 \\
ViSTA-SLAM \cite{zhang2025vista} & \sbest 5.5 && 7.3 & 3.5 & \tbest 2.8 & \best 5.5 & \tbest 12.9 & \best 3.5 & \sbest 2.9 \\

\midrule
HI-SLAM2 \cite{zhang2024hi} & \sbest 5.5 & &\tbest 3.8 & \tbest 3.1 & \sbest 2.6 & \tbest 8.5 & 14.2 & \tbest 4.0 & \best 2.4 \\
\textbf{SING3R-SLAM (Ours)}~~ & \best 4.9 && \best 3.3 & \best 2.7 & 4.5 & \sbest 7.2 & \best 8.7 & \sbest 3.6 & 4.5 \\

\bottomrule
\end{tabular}
}
% \vspace{-2mm}
\caption{\textbf{Quantitative Comparison of Pose Estimation Accuracy on 7-scenes \cite{shotton2013scene} (unit: cm).} Our method achieves the best tracking performance among both 3D reconstruction-based and Gaussian-based approaches, reducing the average ATE by 12\%.}
% \vspace{-3mm}
\label{tab:7scenes}
\end{table*}

\subsubsection{Loop Ddetection}
Given previous views with features $\{f_t\}_{t=0}^{M}$, poses $\{T_t^{\text{world}}\}_{t=0}^{M}$ and depth maps $\{D_t\}_{t=0}^{M}$, and the incoming frame with feature $f_{M+1}$, pose $T_{M+1}^{\text{world}}$ and depth $D_{M+1}$, we evaluate the overlap ratio $r_{q,m}^{\text{overlap}}$ by reprojecting current view to all previous views. In addition, we compute the feature similarity $r_{q,m}^{\text{feat}}$ using Eq.\ref{eq:similarity}, which helps avoid false positives caused by ambiguous geometric structures. The final loop score is computed as $\mathcal{S}_{q,m} = 0.7\, r_{q,m}^{\text{overlap}} + 0.3\, r_{q,m}^{\text{feat}}$, where subscript $q$ denotes the incoming query view and $m$ denotes a previous view.
If the score exceeds a threshold, the incoming frame and its matched views are added as loop candidates, and the temporally farthest match is selected as the loop closure pair.

\subsubsection{Pose Graph Optimization and Gaussian Updates}
Suppose the incoming query frame $C_{i,j}$ is detected to form a loop with matched frame $C_{m,n}$ from a previous submap $G_m$. 
We construct a new submap with two frames $\{C_{\text{loop},j}\} \in G_{\text{loop}}$, where $C_{\text{loop},0} = C_{i,j}$ and $C_{\text{loop},1} = C_{m,n}$. 
Using the monocular geometry estimator, we predict the local poses and then transform them into world coordinates using Eq.\ref{eq:submap_align} via the overlapping frame $C_{i,j}$, yielding $T_{\text{loop},0}^{\text{world}}, T_{\text{loop},1}^{\text{world}}$. 
We then apply a submap-level pose graph optimization to close the loop:
% \begin{equation}
\begin{align}
\min_{\mathcal{T}} \; \sum_{t=0} \Big(
    & \big\| \mathcal{T}_{t-1}T_{t-1,K}^{\text{world}} - \mathcal{T}_{t}T_{t,0}^{\text{world}} \big\| \Big) + \notag \\
    &  \big\| \mathcal{T}_iT_{i,j}^{\text{world}} - \mathcal{T}_{\text{loop}}T_{\text{loop},0}^{\text{world}} \big\| + \notag \\
    &  \big\| \mathcal{T}_{\text{loop}}T_{\text{loop},1}^{\text{world}} - \mathcal{T}_mT_{m,n}^{\text{world}} \big\|
,
\label{eq:loop_closure}
\end{align}
% \end{equation}
where $\mathcal{T} \in SE(3)$ denotes the rigid transformations applied on each submap. The first term enforces consistency between adjacent submaps, while the last term ensures the closure of the detected loop.

Once the submap transformations are optimized, we update the global Gaussian map by applying each transformation to the Gaussians according to the submap ID from which they were generated:
\begin{equation}
\mu_{t, k}' = \mathcal{T}_i\mu_{t, k}, \quad \ R_{t, k}' = \mathcal{T}_tR_{t, k}, \quad \forall k,
\end{equation}
where $t$ represents the corresponding $t$-th submap.

\subsubsection{Global Bundle Adjustment}
While submap-level pose graph optimization can correct loop closure errors, it lacks geometric constraints and therefore cannot achieve highly precise global alignment. By leveraging the Gaussian map, we further perform off-line joint optimization across the entire sequence using Eq.\ref{eq:map_update} to refine global geometric and poses. 
After optimization, the refined poses and depth maps are reprojected and fused to generate updated point clouds and meshes for evaluation, while the Gaussian map provides a compact representation suitable for downstream tasks.

\begin{figure*}[ht]
\includegraphics[width=17.5cm]{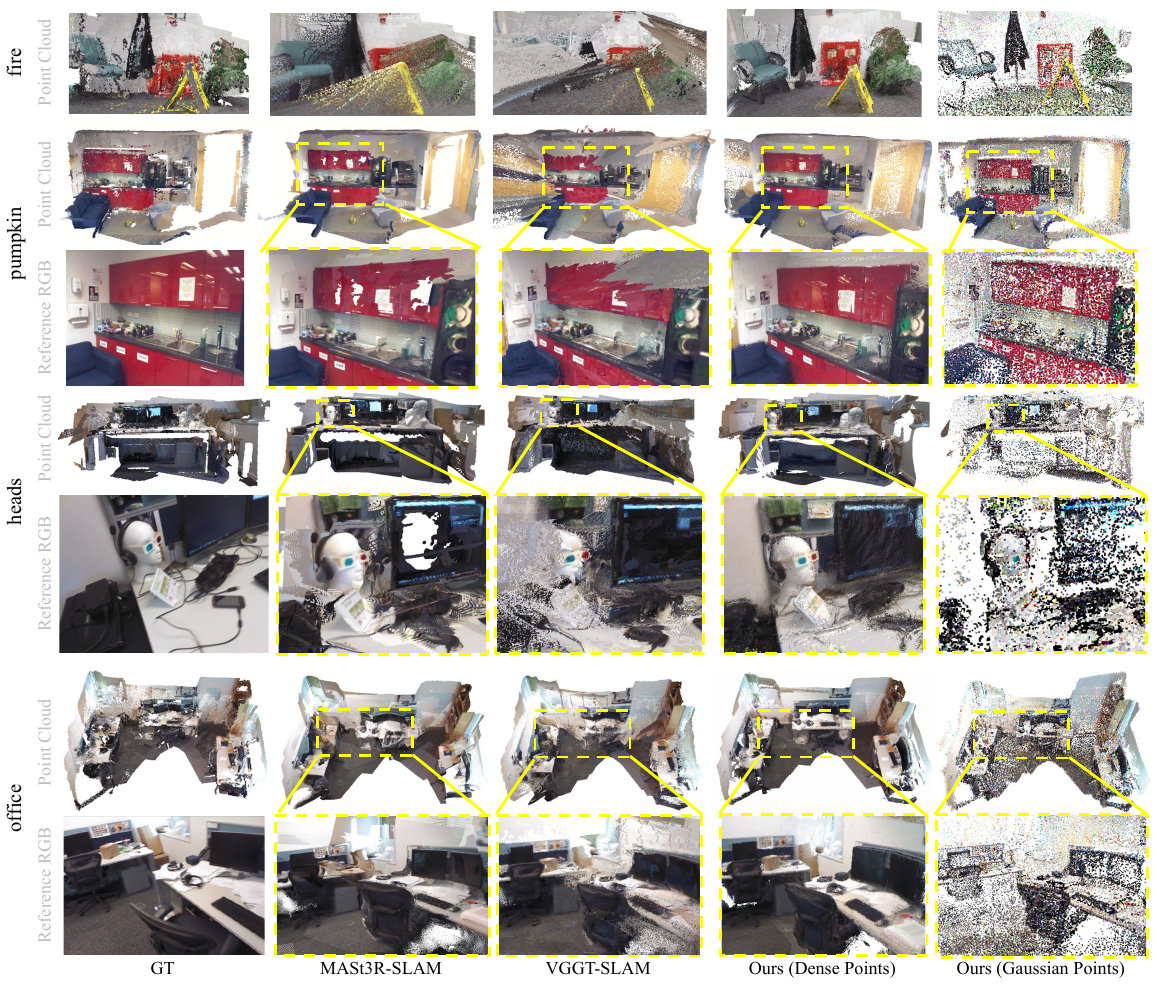}
\centering

\caption{\textbf{Qualitative Comparison of Reconstructed Point Clouds on 7-scenes \cite{shotton2013scene}.} We show the reconstructed point clouds with zoomed-in views for all methods. We also show the compact Gaussian map in the last column. While other 3D methods produce many redundant points which degrade visual quality, our reconstruction preserves geometric structures more accurately.
}
\vspace{-5mm}
\label{fig:7scenes}
\end{figure*}

\section{Experiments}
\label{sec:experiments}
\subsection{Experimental Settings} 
\subsubsection{Datasets}
We comprehensively evaluate our method on two public datasets: 7-scenes \cite{shotton2013scene}, ScanNet-v2 \cite{dai2017scannet}.

\subsubsection{Baselines}
We compare our proposal with several state-of-the-art methods. Among 3D reconstruction-based SLAM methods, we compare against MASt3R-SLAM \cite{murai2025mast3r}, VGGT-SLAM \cite{maggio2025vggt} and ViSTA-SLAM \cite{zhang2025vista}. For pose estimation, we additionally include CUT3R \cite{wang2025continuous}. As for Gaussian-based SLAM methods, we compare against MonoGS \cite{matsuki2024gaussian}, Splat-SLAM \cite{sandstrom2025splat}, and HI-SLAM2 \cite{zhang2024hi}. Finally, we compare with two RGB-D methods, SplaTAM \cite{keetha2024splatam} and Gaussian-SLAM \cite{yugay2023gaussian}, in terms of rendering quality.

\subsubsection{Metrics} 
To evaluate our method, we follow standard practice and report surface accuracy using Accuracy, Completeness, and Chamfer Distance. Visual fidelity of synthesized novel views is measured with PSNR, SSIM, and LPIPS~\cite{zhang2018unreasonable}, while Absolute Trajectory Error (ATE) is used for pose evaluation.

\subsubsection{Implementation} CUT3R \cite{wang2025continuous} is used as our monocular geometry estimator $\mathcal{D}$ and RaDe-GS \cite{zhang2024rade} is used as our Gaussian rasterizer. We train on a single NVIDIA RTX 4090 GPU. We choose $K=6$ frames for each submap. For keyframe selection, we set the default threshold as 0.7. For loop detection, we set the threshold as 0.5. For submap registration, we set $\lambda_{scaleD}=10$. For local bundle adjustment, we apply online optimization with 20 iterations and set the sliding window $W=10$ and $\lambda_{D}=10, \lambda_{N}=0,  \lambda_{DN}=0.01, \lambda_{S}=10$. For global bundle adjustment, we set $\lambda_{D}=0, \lambda_{N}=0.005,  \lambda_{DN}=0.005, \lambda_{S}=0$ and apply off-line optimization with 20000 iterations.

\begin{table}[t]
\centering
\resizebox{0.9\linewidth}{!}{
\begin{tabular}{lccc}
\toprule
Method & Acc. $\downarrow$ & Complet. $\downarrow$ & Chamfer $\downarrow$ \\
\midrule
DROID-SLAM \cite{teed2021droid} & 0.141 & \tbest 0.048 & 0.094 \\
% Spann3R \cite{wang20253d}& 0.069 & \sbest 0.047 & 0.058 \\
MASt3R-SLAM \cite{murai2025mast3r}& \tbest 0.068 & \best 0.045 & \tbest 0.056 \\
VGGT-SLAM \cite{maggio2025vggt}& \sbest 0.052 & 0.058 & \sbest 0.055 \\
\textbf{SING3R-SLAM (Ours)} & \best 0.045 & \sbest 0.047 & \best 0.046 \\
\bottomrule
\end{tabular}}
% \vspace{-3mm}
\caption{\textbf{Quantitative Comparison of  3D Reconstruction on 7-Scenes \cite{shotton2013scene} (unit: m).} Our method achieves state-of-the-art performance on 3D reconstruction. Our approach remains robust and consistently outperforms existing methods, further supported by strong qualitative comparisons.}
% \vspace{-3mm}
\label{tab:recon_7scenes}
\end{table}

\subsection{Geometry Reconstruction}
We evaluate 3D reconstruction quality on the 7-Scenes dataset~\cite{shotton2013scene}, as shown in Table~\ref{tab:recon_7scenes}. 
Our method achieves state-of-the-art performance in 3D reconstruction, indicating more precise and well-aligned geometry. Although our approach does not rely on ground-truth depth and instead incorporates normal regularization to promote smooth reconstructions, we observe that at the boundaries of large planar regions (e.g., walls), the lack of direct supervision can lead to slight over-smoothing, causing minor deviations from ideal geometry. Nevertheless, despite these limitations, our method still delivers significant overall improvements, demonstrating its robustness and effectiveness.
Qualitatively, our method consistently produces accurate and complete geometry. As shown in Fig.~\ref{fig:7scenes} (\textit{pumpkin}), MASt3R-SLAM~\cite{murai2025mast3r} fails to reconstruct reflective cabinet doors, while VGGT-SLAM~\cite{maggio2025vggt} introduces structural artifacts. In contrast, our approach maintains a global Gaussian map and leverages multi-view consistency to suppress floating artifacts commonly observed in monocular depth reconstruction (e.g., the scattered points in MASt3R-SLAM and VGGT-SLAM). As a result, our reconstructions are both more geometrically faithful and significantly improved overall.
As shown in the bottom-left of Fig.~\ref{fig:teaser}, VGGT-SLAM~\cite{maggio2025vggt} reconstructs the same wall inconsistently from different viewpoints, causing misaligned surfaces. In contrast, our global Gaussian map enforces multi-view geometric consistency, yielding complete and well-aligned wall reconstructions. Additional video results are provided in the supp. mat..

\begin{figure*}[t]
\includegraphics[width=15cm]{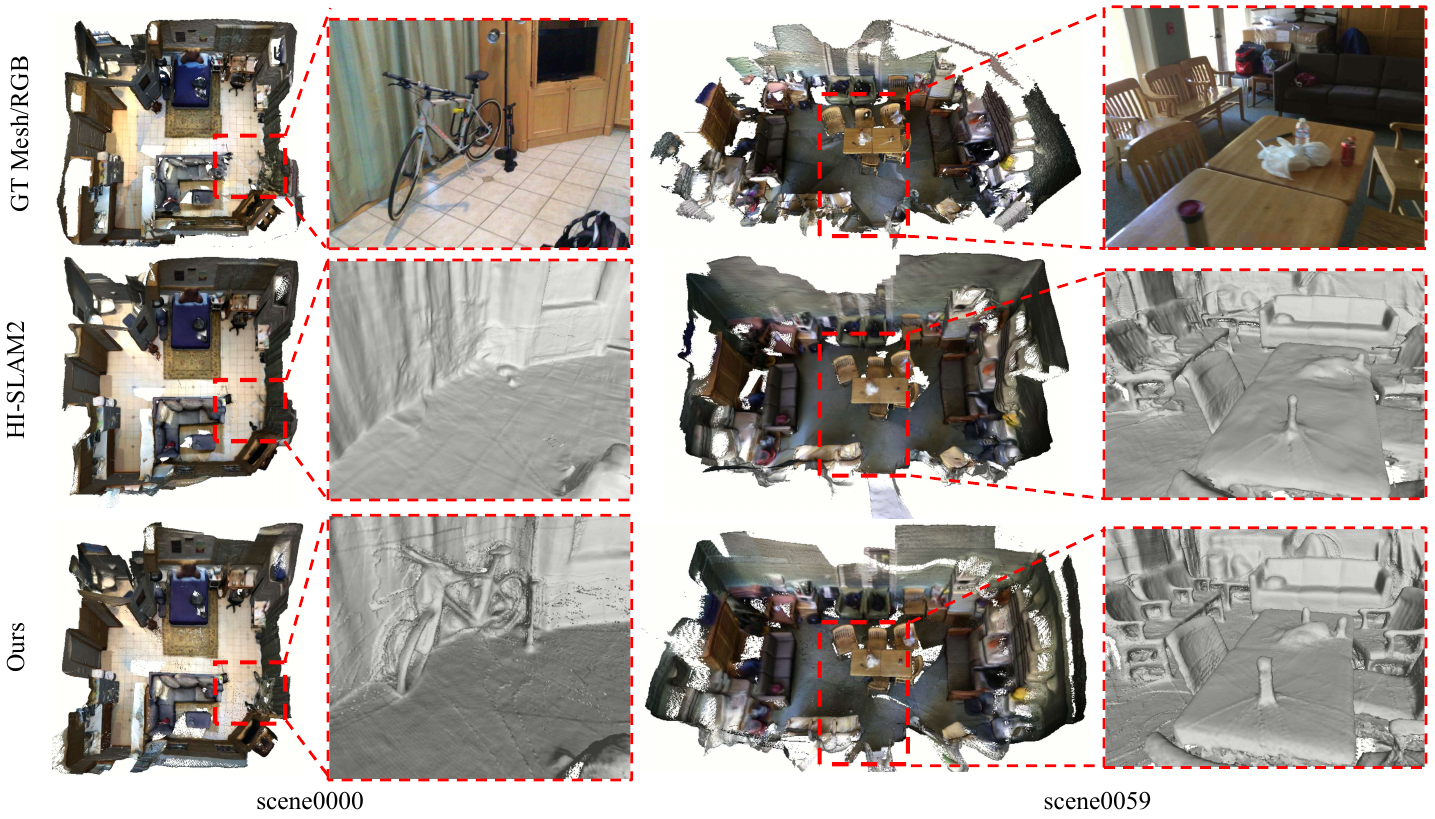}
\centering
% \vspace{-2mm}
\caption{\textbf{Qualitative Comparison of Reconstructed Meshes on Scannet-v2 \cite{dai2017scannet}.} We compare our reconstructed meshes with HI-SLAM2 \cite{zhang2024hi}. Our method successfully captures fine scene details, such as the bicycle in scene \textit{0000} and the chair’s armrests in scene \textit{0059}, demonstrating superior geometric fidelity and reconstruction quality.
}
\label{fig:scannet}
\vspace{-2mm}
\end{figure*}

We also visualize the reconstructed meshes in Fig.~\ref{fig:scannet} to demonstrate the versatility of our representation. Compared with HI-SLAM2~\cite{zhang2024hi}, which relies on multiple modules to separately estimate camera poses and scene geometry, our approach directly reconstructs submap-level local geometry using a 3D reconstruction prior, leading to richer local details. These local reconstructions are then integrated into a global Gaussian map, which preserves fine structures while maintaining global consistency. As a result, our method better retains small-scale details, such as the bicycles in ScanNet-v2 scene\_\textit{0000}.

\subsection{Pose Estimation}
We evaluate the camera pose estimation accuracy of our system against both 3D reconstruction-based and Gaussian-based approaches. As reported in Table~\ref{tab:7scenes}, SING3R-SLAM achieves the best average pose estimation performance on the 7-scenes dataset, demonstrating a substantial improvement of over 10\% in average ATE compared to existing methods. The performance gain is particularly pronounced on challenging scenes such as \textit{pumpkin}, where our approach significantly outperforms all baselines. On Scannet-v2 dataset, as shown in Table~\ref{tab:scannet}, our method also achieves competitive results. These results highlight the effectiveness of integrating local submaps with a globally optimized Gaussian map, which not only provides a unified global map but also ensures robust pose estimation across diverse and complex indoor environments.

\begin{table}[t]
\centering
\resizebox{1.\linewidth}{!}{
\begin{tabular}{lccccccccc}
\hline
Method && Avg. && 00 & 59 & 106 & 169 & 181 & 207 \\
\hline
MASt3R-SLAM \cite{murai2025mast3r} && 7.95 && 6.96 & \tbest 8.48 & 9.53 & 8.34 & \best 7.16 & \best 7.20 \\
VGGT-SLAM \cite{maggio2025vggt} && 19.3 && 12.79 & 10.21 & 39.89 & 22.38 & 12.63 & 17.97 \\
\hline
MonoGS \cite{matsuki2024gaussian} && 122.7 && 149.2 & 96.8 & 155.5 & 140.3 & 92.6 & 101.9 \\
Splat-SLAM \cite{sandstrom2025splat} && \tbest 7.66 && \best 5.57 & 9.11 & \tbest 7.09 & \tbest 8.26 & \tbest 8.39 & \tbest 7.53 \\
HI-SLAM2 \cite{zhang2024hi}  && \best 7.16 && \tbest 5.82 & \sbest 7.30 & \sbest 6.80 & \sbest 8.25 & \sbest 7.41 & \sbest 7.40 \\
\textbf{SING3R-SLAM (Ours)}   && \sbest 7.41 && \sbest 5.70 & \best 7.20 & \best 6.75 & \best 8.02 & 8.47 & 8.31 \\
\hline
\end{tabular}
}
% \vspace{-3mm}
\caption{\textbf{Quantitative Comparison of Pose Estimation Accuracy on ScanNet-v2~\cite{dai2017scannet} (unit: cm).} SING3R-SLAM achieves competitive trajectory accuracy compared with state-of-the-art SLAM systems.}
\label{tab:scannet}
\end{table}

\begin{table}[t]
\centering
\resizebox{0.4\textwidth}{!}{
\begin{tabular}{lccc}
\toprule
Method & PSNR $\uparrow$ & SSIM $\uparrow$ & LPIPS $\downarrow$ \\
\midrule
\multicolumn{4}{c}{\textbf{RGB-D input}} \\
\midrule
SplaTAM \cite{keetha2024splatam} & 20.42 & 0.78 & 0.38 \\
Gaussian-SLAM \cite{yugay2023gaussian} & 27.67 & \best 0.92 & 0.25 \\
\midrule
\multicolumn{4}{c}{\textbf{RGB input}} \\
\midrule
Splat-SLAM \cite{sandstrom2025splat} & \sbest 29.48 & 0.85 & \best 0.18 \\
HI-SLAM2 \cite{zhang2024hi} & \tbest 29.27 & \tbest 0.88 & \tbest 0.24 \\
\textbf{SING3R-SLAM (Ours)} & \best 30.47 & \sbest 0.89 & \sbest 0.21 \\
\bottomrule
\end{tabular}}
% \vspace{-2mm}
\caption{\textbf{Quantitative Comparison of Average Rendering Performance on ScanNet-v2 \cite{dai2017scannet}.}}
\vspace{-5mm}
\label{tab:render_scannet}
\end{table}

\subsection{Rendering}
By adopting a Gaussian map as the scene representation, our system naturally supports novel view synthesis (NVS) through rendering. 
NVS has recently become an important metric for evaluating 3D reconstruction quality, as it reflects both geometric accuracy and multi-view consistency. 
As shown in Table~\ref{tab:render_scannet}, our method achieves the best performance among both RGB-D and RGB-only Gaussian SLAM methods. 
We believe this gain comes from our submap-level loop closure and per-frame refinement, which significantly improve global alignment and geometric consistency.

\begin{table}[t]
\centering
  \resizebox{0.4\textwidth}{!}{
  \begin{tabular}{cccccccc}
    \toprule
    \# & frame refine & LBA & GBA & Loop & ATE $\downarrow$ & PSNR $\uparrow$ \\
    \cmidrule(lr){1-1} \cmidrule(lr){2-5} \cmidrule(lr){6-7} 
    (a) & \ & \ & \ & \ & 104.15 & - \\
    (b) & \ & \ & \ & \checkmark & 34.25 & - \\
    (c) & \ & \checkmark & \checkmark & \checkmark & 12.25 & 25.66 \\
    (d) & \checkmark & \checkmark & \ & \checkmark & 9.39 & 26.43 \\
    (e) & \checkmark & \checkmark & \checkmark & \ & 24.54  & 20.17 \\
    (f) & \checkmark & \checkmark & \checkmark & \checkmark & \textbf{7.20} & \textbf{29.44} \\
    \bottomrule
  \end{tabular}
  }
  % \vspace{-0.2cm}
  \caption{\textbf{Ablations on Key Components.}}
  \label{table:ablation}
  % \vspace{-0.1cm}
\end{table}

\begin{table}[h]
\centering
\resizebox{0.9\linewidth}{!}{
\begin{tabular}{lcccc}
\toprule
Method & Tracking & LBA & GBA & Map Size \\
\midrule
MASt3R-SLAM \cite{murai2025mast3r}& \textbf{5min} & - & - & 110 MB \\
HI-SLAM2 \cite{zhang2024hi}& 8min & 12min & 10min & 9 MB \\
\textbf{SING3R-SLAM (Ours)} & \textbf{5min} & \textbf{10min} & \textbf{8min} & \textbf{7 MB} \\
\bottomrule
\end{tabular}}
% \vspace{-2mm}
\caption{\textbf{Performance Analysis.} SING3R-SLAM focuses on map reconstruction, with camera poses jointly optimized rather than provide a separate tracking module. For fair comparison, we consider submap local reconstruction as the equivalent of tracking in conventional SLAM.}
% \vspace{-3mm}
\label{tab:runtime}
\end{table}

\subsection{Ablations and Performance Analysis}
All ablations and performance analysis are conducted on the scene\_\textit{0059} of the ScanNet-v2 \cite{dai2017scannet} dataset.

\subsubsection{Ablations}
In Table~\ref{table:ablation}, we evaluate the contributions of the key components in our system, including per-frame refinement, local bundle adjustment, global bundle adjustment, and loop closure. The last row (f) reports the results of the full model.

We first analyze the performance without the global Gaussian map and loop closure. Note that directly applying CUT3R to long sequences often fails due to accumulated drift. Row (a) shows the result using only the CUT3R module with submap-based reconstruction, where the ATE indicates failure on long sequences. Row (b) adds our submap-level loop closure, which significantly improves convergence compared with row (a), demonstrating the effectiveness of the proposed loop detection and correction strategy.

We then examine the contributions of the Gaussian map optimization components. Comparing row (c) with the full model in row (f), we observe that frame-level refinement is essential, leading to substantial improvements in both ATE and rendering quality. Comparing row (d) with row (f) further shows that offline global bundle adjustment provides additional gains in accuracy. Finally, the comparison among rows (d), (e), and (f) demonstrates that both the proposed loop closure and global bundle adjustment are necessary to achieve the best performance.

\subsubsection{Performance Analisys}
Table~\ref{tab:runtime} reports runtime and memory usage on scene\_\textit{0059} of ScanNet-v2 \cite{dai2017scannet}. 
SING3R-SLAM reconstructs a globally consistent map with camera poses jointly optimized during reconstruction rather than as a separate tracking module. 
For fair comparison, we consider submap-level local reconstruction (Sec.~\ref{sec:tracker}) as the equivalent of tracking in conventional SLAM. SING3R-SLAM achieves tracking times comparable to MASt3R-SLAM, and our global bundle adjustment module can run off-line in parallel, minimizing overhead.
MASt3R-SLAM relies on per-frame, per-pixel point maps, resulting in very large map sizes and performing only tracking without map optimization. 
In contrast, our global Gaussian map is compact (7 MB) yet enables dense reconstruction via per-pixel depth rendering. 

\section{Conclusion}
\label{sec:conclusion}
We present SING3R-SLAM, a monocular indoor SLAM system that integrates local 3D reconstruction priors with a globally consistent Gaussian map. By aligning local submaps into the global map and applying differentiable bundle adjustment via rendering, our method jointly refines camera poses and scene geometry without complex feature matching. This produces a compact, globally consistent scene representation that supports dense reconstruction and tasks like novel view synthesis. Experiments on real-world datasets show that SING3R-SLAM achieves excellent tracking and reconstruction performance.

{
    \bibliographystyle{IEEEtran}
    \bibliography{root}
}

\end{document}